\documentclass[preprint,12pt]{elsarticle}

\usepackage[T1]{fontenc}
\usepackage{lmodern}
\usepackage{amsmath,amssymb,bm}
\usepackage{graphicx}
\usepackage{booktabs}
\usepackage{multirow}
\usepackage{tabularx}
\usepackage{array}
\usepackage{makecell}
\usepackage{threeparttable}
\usepackage{microtype}
\usepackage{url}
\usepackage{xcolor}
\usepackage{placeins}

\graphicspath{{figures/}{./}}

\newcolumntype{Y}{>{\centering\arraybackslash}X}
\newcommand{\cmark}{\checkmark}

\journal{Results in Engineering}
\biboptions{sort&compress}

\begin{document}

\begin{frontmatter}

\title{Beyond Accuracy: Quantifying Pulmonary Attribution in Anatomy-Guided Chest X-Ray Classification Under Domain Shift}

\author[ruet]{Abdullah Al Mamun}
\author[ruet]{Md. Nasif Osman Khansur}
\author[ruet]{Md Ashraful Hossen Akash}
\author[erl]{Md. Kishor Morol}
\author[mmu1,mmu2]{Tze Hui Liew\corref{cor1}}
\ead{thliew@mmu.edu.my}

\cortext[cor1]{Corresponding author.}
\address[ruet]{Department of Computer Science and Engineering,
Rajshahi University of Engineering \& Technology (RUET),
Kazla, Rajshahi 6204, Bangladesh}
\address[erl]{Elite Research Lab LLC, New York, USA}
\address[mmu1]{Faculty of Information Science and Technology, Multimedia University, Melaka, Malaysia}
\address[mmu2]{Centre for Intelligent Cloud Computing (CICC), COE of Advanced Cloud, Faculty of Information Science \& Technology, Multimedia University, 75450 Melaka, Malaysia}

\begin{abstract}
Deep-learning models can achieve strong chest X-ray (CXR) classification performance without establishing whether their predictions predominantly rely on pulmonary image content. This study evaluates pulmonary attribution containment as an anatomy-related reliability property distinct from diagnostic performance. We propose DBCA-SegNet-MGAP, a multi-task anatomy-guided CNN--Transformer framework that combines complementary feature representations through bidirectional cross-backbone attention, predicts a soft lung mask, and incorporates this anatomical prior directly into classification through Mask-Guided Adaptive Global Average Pooling (MGAP). Pulmonary attribution containment is quantified using the Anatomical Local Energy Ratio (ALR) and high-intensity cumulative ALR (cALR@0.9). Experiments were repeated across three training seeds using the COVID-19 Radiography Database for four-class internal testing and a locked Shenzhen-to-Montgomery protocol for zero-shot external tuberculosis testing. On COVID-19, the proposed model achieved a weighted F1 of $0.9615 \pm 0.0015$ and macro ROC-AUC of $0.9906 \pm 0.0007$. In an architecture-matched dual-bridge comparison, replacing conventional GAP with MGAP increased ALR from $0.3878 \pm 0.0098$ to $0.7086 \pm 0.0104$ and cALR@0.9 from $0.5265 \pm 0.0101$ to $0.9905 \pm 0.0018$, while weighted F1 remained essentially unchanged ($0.9618 \pm 0.0015$ vs.\ $0.9615 \pm 0.0015$). Under locked external transfer to Montgomery, ROC-AUC remained $0.9080 \pm 0.0043$ and pulmonary ALR remained $0.6466 \pm 0.0081$, whereas weighted F1 decreased to $0.7528 \pm 0.0080$ and ECE increased to $0.1683 \pm 0.0055$. These findings show that diagnostic discrimination, calibration, and pulmonary attribution containment are distinct model properties and support their joint evaluation under internal testing and external domain shift.
\end{abstract}


\end{frontmatter}

% ============================================================
\section{Introduction}
\label{sec:introduction}
% ============================================================

Chest radiography is a widely used imaging modality for the assessment of pulmonary and thoracic abnormalities because it is rapid, comparatively inexpensive, and broadly accessible. Deep-learning methods have consequently been investigated extensively for automated chest X-ray (CXR) classification, including pneumonia, tuberculosis (TB), COVID-19, pulmonary opacity, and other thoracic conditions. CNN-based, Transformer-based, and more recent hybrid architectures have achieved strong diagnostic performance on public CXR datasets \cite{zhang2021pmgan,ctranscnn,saporta2022saliency}. However, predictive performance alone does not establish whether a model has learned a diagnostically meaningful decision strategy.

This distinction is particularly important in chest radiography because a CXR contains substantial information outside the pulmonary field. Image borders, embedded text, acquisition protocol, positioning, scanner characteristics, preprocessing patterns, and dataset-specific signatures may correlate with diagnostic labels. A classifier optimized only for label prediction is not required to distinguish pulmonary evidence from such correlations. Previous work has shown that radiographic classifiers can exploit non-clinical shortcuts while maintaining apparently strong internal performance \cite{degrave2021shortcuts}. Consequently, high accuracy or ROC-AUC cannot by itself establish that a model's decision-related attribution is concentrated within clinically relevant pulmonary anatomy.

Anatomical guidance provides one approach to reducing this disconnect. Existing studies have incorporated lung or anatomical part information through cropping, auxiliary segmentation, part-aware representation learning, mask-guided attention, and feature refinement \cite{zhang2021pmgan,agfrnet,wuAnatomyGuided}. These approaches demonstrate that anatomical priors can influence thoracic image classification, but they also expose an important distinction: learning pulmonary anatomy is not necessarily equivalent to using pulmonary anatomy during classification. An auxiliary network may predict the lung field accurately while the classification head remains free to aggregate discriminative information from non-pulmonary regions. Establishing anatomical prediction alone is therefore insufficient to demonstrate anatomically concentrated classification behavior.

The feature-aggregation stage provides a direct point at which this distinction can be investigated. Conventional Global Average Pooling (GAP) aggregates spatial features uniformly and does not explicitly distinguish pulmonary from extrapulmonary locations. A learned pulmonary prior can instead be incorporated directly into the aggregation rule. In this study, Mask-Guided Adaptive Global Average Pooling (MGAP) uses a jointly predicted soft lung mask to continuously weight the final spatial representation before classification. Unlike hard lung cropping, the complete radiograph remains available to the feature encoders; the predicted pulmonary probability modifies the relative contribution of spatial features during normalized pooling rather than removing image regions outright.

The representation supplied to such an aggregation mechanism may also influence its effect. CNNs provide strong inductive biases for local texture and morphology, whereas hierarchical vision Transformers offer complementary contextual modeling. Hybrid CNN--Transformer systems have therefore become increasingly common, including approaches based on direct feature interaction, multiscale cross-attention, and lesion-aware Transformer representations \cite{ctranscnn,shiCrossAttention,shinThoraxformer,fuHybridTransformer}. Accordingly, CNN--Transformer fusion or cross-attention alone is not treated here as the central novelty. Instead, the present study uses Bidirectional Cross-Backbone Attention Blocks (BCABs) to construct controlled local--global representation states and asks whether the effect of anatomical pooling changes with the preceding feature representation.

A second challenge concerns how anatomical behavior should be evaluated. Grad-CAM \cite{selvaraju2017gradcam}, Layer-CAM \cite{jiang2021layercam}, and related methods are widely used to visualize class-associated image regions, but selected heatmaps provide limited evidence about model behavior across an entire test cohort. Moreover, visually plausible saliency does not necessarily imply accurate localization or faithful explanation \cite{saporta2022saliency}. The present study therefore separates qualitative visualization from quantitative pulmonary attribution containment. The Anatomical Local Energy Ratio (ALR) measures the fraction of total predicted-class attribution mass contained within the lung field, whereas cumulative ALR at a high attribution threshold (cALR@0.9) evaluates whether the strongest retained attribution is also pulmonary. These measures are deliberately interpreted as attribution containment rather than lesion localization, causal explanation, or diagnostic correctness.

Reliability must additionally be examined beyond the development distribution. Deep-learning systems that perform strongly during internal testing often deteriorate when evaluated on external cohorts because patient populations, acquisition equipment, disease prevalence, imaging protocols, and other data characteristics may differ \cite{yu2022external}. Furthermore, different reliability properties need not deteriorate equally. ROC-AUC reflects discrimination or case ranking, whereas fixed-threshold metrics depend on the selected operating point, and probability calibration describes whether predicted confidence remains numerically meaningful \cite{guo2017calibration}. A model may therefore retain useful external discrimination while showing poorer calibration or threshold-dependent classification performance. Current reporting guidance for medical-imaging AI consequently emphasizes explicit separation of internal and external testing and transparent description of data partitions and reference standards \cite{claim2024}.

\begin{figure}[!t]
    \centering
    \includegraphics[width=\textwidth]{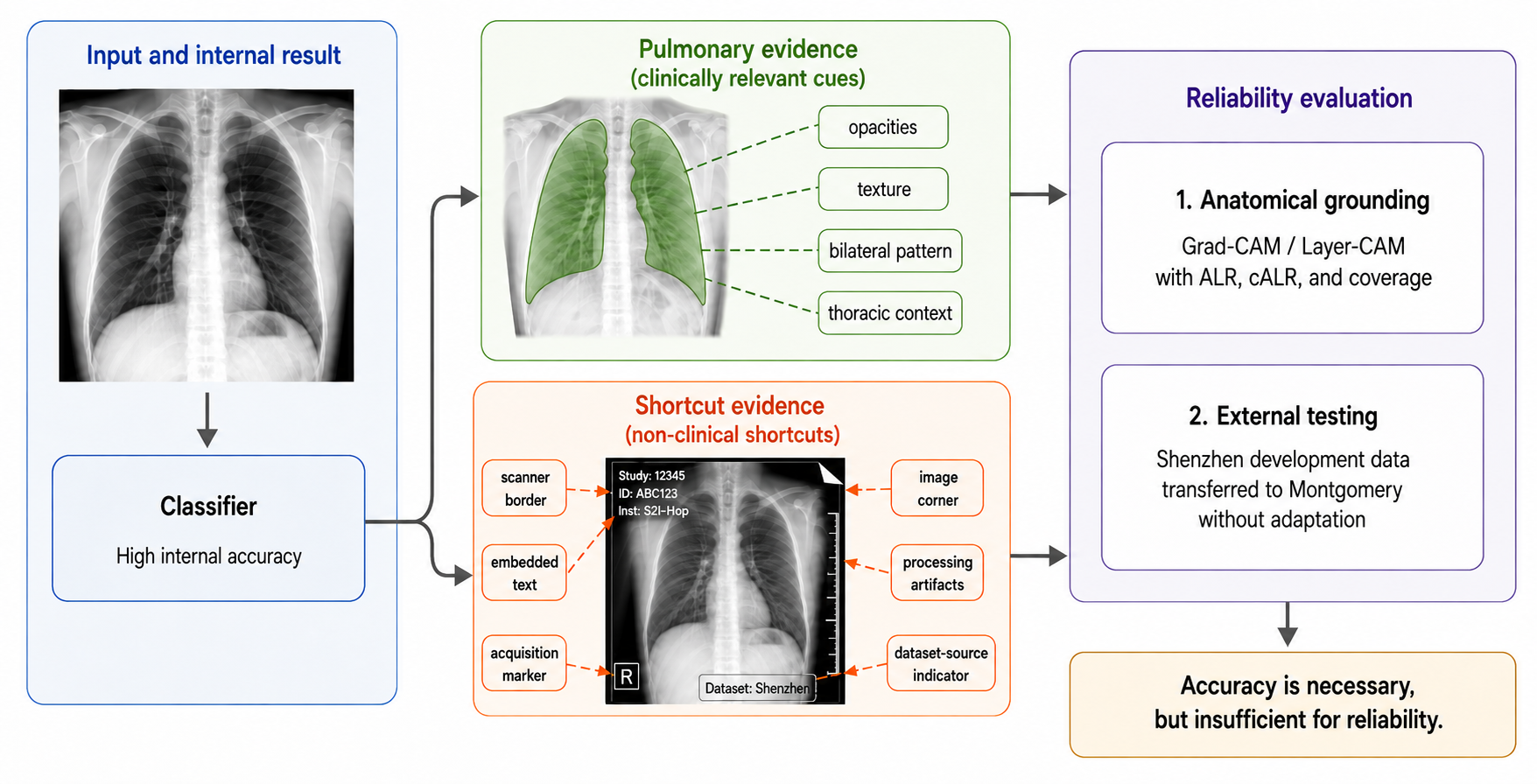}
    \caption{Reliability gap in chest X-ray classification: diagnostic performance
    versus pulmonary attribution containment and external robustness. Strong
    internal diagnostic performance does not establish whether class-related
    attribution is concentrated within the pulmonary field or whether diagnostic
    behavior is preserved under external cohort shift. The pulmonary anatomical
    prior represents the target anatomical region used for anatomy-guided feature
    aggregation. Potential non-pulmonary cues are shown as conceptual examples and
    are not presented as confirmed shortcuts used by the trained models.}
    \label{fig:reliability_gap}
\end{figure}

Figure~\ref{fig:reliability_gap} summarizes the reliability gap motivating the study. The potential non-pulmonary cues shown in the diagram are conceptual shortcut risks rather than confirmed features used by the trained models.

These considerations motivate a \emph{beyond-accuracy} perspective in which diagnostic competence, pulmonary attribution containment, calibration, and external robustness are treated as related but non-equivalent model properties. To investigate these relationships, we develop DBCA-SegNet-MGAP, a multi-task anatomy-guided CNN--Transformer framework. ResNet-50 and Swin-Tiny provide complementary convolutional and hierarchical Transformer representations, while BCABs permit reciprocal feature exchange at intermediate resolutions. A progressive decoder jointly predicts a continuous lung probability map, which is used both for auxiliary anatomical supervision and as the spatial prior for MGAP.

The experimental design is structured to isolate these effects. Conventional CNN and Transformer classifiers provide diagnostic reference baselines, while a lightweight architecture serves as a capacity control. For the full-capacity framework, late-fusion, single-bridge, and dual-bridge representations are each evaluated with GAP and MGAP, enabling architecture-matched GAP$\rightarrow$MGAP comparisons without changing the preceding representation. Lung segmentation and classifier attribution are evaluated separately, and principal comparisons are repeated across three independent training seeds.

External reliability is evaluated through a locked Shenzhen-to-Montgomery TB protocol. Model development and checkpoint selection are performed using Shenzhen only. The selected checkpoint is then applied to Montgomery without target-domain fine-tuning, recalibration, checkpoint reselection, or threshold optimization. This permits discrimination, fixed-threshold performance, calibration, and pulmonary attribution to be examined independently under cohort shift.

The study makes four connected contributions:
\begin{enumerate}
    \item a jointly predicted soft pulmonary prior is incorporated directly into normalized feature aggregation through MGAP rather than remaining solely an auxiliary segmentation output or being converted into a hard crop;
    \item quantitative pulmonary-attribution metrics and architecture-matched GAP/MGAP comparisons isolate the anatomical aggregation intervention from changes in the preceding representation;
    \item lung-segmentation quality is explicitly separated from pulmonary classifier attribution, testing whether learning pulmonary anatomy and using it for classification are equivalent properties; and
    \item diagnostic performance, calibration, and pulmonary attribution are evaluated separately under a locked external domain shift, with a lightweight capacity-control configuration testing whether strong pulmonary containment alone is sufficient for robust transfer.
\end{enumerate}

Accordingly, the study addresses the following research questions:

\noindent\textbf{RQ1.} Can classifiers with comparable diagnostic performance exhibit substantially different pulmonary-attribution distributions?

\noindent\textbf{RQ2.} Can mask-guided feature aggregation increase pulmonary attribution containment without materially compromising diagnostic competence, and does this effect depend on the preceding feature representation?

\noindent\textbf{RQ3.} Are diagnostic discrimination, probability calibration, and pulmonary attribution containment preserved to the same extent under external cohort shift?

The objective is not to establish universal superiority of a particular backbone, attention block, or pooling architecture. Instead, the study examines whether diagnostic competence, pulmonary attribution containment, calibration, and external robustness can vary independently, and whether explicit anatomical feature aggregation can make pulmonary attribution more directly measurable and controllable.

% ============================================================
\section{Related Work}
\label{sec:related}
% ============================================================

\subsection{Deep learning and hybrid representation learning for CXR classification}

Deep-learning approaches to CXR analysis have progressed from predominantly convolutional architectures toward Transformer-based and hybrid representation learning. CNNs such as ResNet \cite{he2016resnet}, DenseNet \cite{huang2017densenet}, EfficientNet \cite{tan2019efficientnet}, and ConvNeXt \cite{liu2022convnext} remain relevant baselines because hierarchical convolution provides an effective inductive bias for localized radiographic texture and morphology. Hierarchical vision Transformers provide a complementary mechanism for content-dependent contextual modeling; Swin Transformer \cite{liu2021swin} combines multiscale feature extraction with shifted-window self-attention.

Hybrid CNN--Transformer methods seek to exploit these complementary properties. CTransCNN \cite{ctranscnn} introduced direct cross-representation interaction for medical-image classification, while more recent CXR systems have investigated multiscale cross-attention, lesion-aware hybrid representations, and explainable hybrid Transformers \cite{shiCrossAttention,shinThoraxformer,fuHybridTransformer}. These studies establish that hybridization and cross-attention are active and increasingly mature research directions. The present work therefore does not position cross-attention alone as the main contribution; instead, BCABs provide controlled representation states on which the effect of anatomical aggregation can be tested.

\subsection{Anatomy-guided and segmentation-assisted classification}

Anatomical information has been incorporated into thoracic classification through lung cropping, part-aware representations, auxiliary segmentation, spatial attention, and anatomy-guided feature refinement. Zhang et al.\ \cite{zhang2021pmgan} introduced part-aware mask-guided attention to integrate anatomical region information into thoracic disease classification. More recent anatomy-guided approaches have incorporated anatomical priors through progressive feature refinement and segmentation-derived localization mechanisms \cite{agfrnet,wuAnatomyGuided}. These studies establish that pulmonary priors can influence classification behavior.

However, successful lung segmentation does not necessarily establish that the final classification head relies strongly on pulmonary features. An auxiliary decoder can learn a high-quality anatomical task while the classifier continues to aggregate information from pulmonary and extrapulmonary spatial locations. The present work targets this distinction by inserting a predicted soft lung prior directly into normalized spatial aggregation and comparing GAP and MGAP within otherwise matched representations.

\subsection{Explainability, shortcut learning, and quantitative pulmonary attribution}

Grad-CAM \cite{selvaraju2017gradcam} and Layer-CAM \cite{jiang2021layercam} are widely used to inspect class-associated spatial regions. Recent CXR studies similarly combine diagnostic models with visual explanation \cite{pediapulmodx,kaushikXAI,fuHybridTransformer}. Nevertheless, qualitative plausibility is not equivalent to validated localization. Saliency benchmarking in CXR analysis has demonstrated important limitations in localization performance \cite{saporta2022saliency}, while shortcut-learning studies have shown that radiographic classifiers may exploit unintended acquisition- or dataset-associated signals \cite{degrave2021shortcuts}.

These observations motivate cohort-level quantitative containment rather than reliance on selected heatmaps alone. ALR measures the fraction of predicted-class attribution mass inside the pulmonary field, while cALR@0.9 asks whether the strongest retained attribution is also pulmonary. Neither metric is interpreted as lesion localization, explanation faithfulness, or causal reasoning.

\subsection{Domain shift, calibration, and external reliability}

Internal test performance does not guarantee preservation of model behavior across cohorts. External evaluation of radiologic deep-learning systems frequently reveals performance deterioration under changes in patient population, acquisition equipment, protocols, and other site-specific factors \cite{yu2022external}. Reliability under domain shift is also multidimensional: ROC-AUC measures ranking discrimination, threshold-dependent metrics depend on the operating point, and calibration measures whether predicted confidence remains numerically meaningful \cite{guo2017calibration}.

Public Shenzhen and Montgomery CXR cohorts provide a relevant setting for TB transfer analysis \cite{jaeger2014cxr}. Previous work has demonstrated the feasibility of deep-learning-based TB screening and cross-population evaluation \cite{pasa2019tb,dasCrossPopulation}. The present study differs in its explicitly locked source-to-target protocol: Shenzhen is used for model development and checkpoint selection, whereas Montgomery is reserved for external testing without target-domain fine-tuning, recalibration, checkpoint reselection, or threshold optimization.

\subsection{Research gap and study positioning}

Across the literature, anatomical prediction, anatomical use, diagnostic performance, and external reliability are often investigated as separate problems. The unresolved issue addressed here is not simply how to construct another accurate hybrid classifier, but how to connect a learned pulmonary prior directly to classifier aggregation, isolate the effect of that intervention from the preceding representation, quantify pulmonary attribution across a test cohort, and determine whether diagnostic and anatomy-related properties are preserved similarly under cohort shift.

DBCA-SegNet-MGAP is positioned around this experimental gap. The study therefore treats
\begin{align*}
\text{diagnostic competence} 
&\neq \text{pulmonary-prior quality} \\
&\neq \text{pulmonary attribution containment} \\
&\neq \text{external reliability},
\end{align*}
and tests their relationships under controlled internal and locked external evaluation.

% ============================================================
\section{Methodology}
\label{sec:methods}
% ============================================================

\subsection{Study design and experimental framework}

This study employed a retrospective controlled computational design to evaluate CXR classification beyond diagnostic performance alone. Four complementary properties were examined: diagnostic discrimination, probability calibration, pulmonary segmentation, and pulmonary attribution containment. External robustness was additionally evaluated through a locked cross-cohort TB experiment.

Two independent task tracks were used. The first used the COVID-19 Radiography Database for four-class classification of Normal, Lung Opacity, COVID-19, and Viral Pneumonia. The second used Shenzhen for binary Normal-versus-TB development and internal testing, followed by external testing on Montgomery. The task-specific models used separate classifier outputs and checkpoints.

The experimental hierarchy comprised conventional pretrained classifiers as diagnostic baselines, a lightweight configuration as a capacity control, architecture-matched GAP/MGAP conditions for mechanism isolation, and the complete DBCA-SegNet-MGAP framework. Montgomery was treated strictly as an external test cohort and was not used for training, model selection, recalibration, normalization adaptation, or decision-threshold optimization, consistent with the distinction between internal and external testing emphasized in CLAIM \cite{claim2024}.

\begin{figure}[!t]
    \centering
    \includegraphics[width=\textwidth]{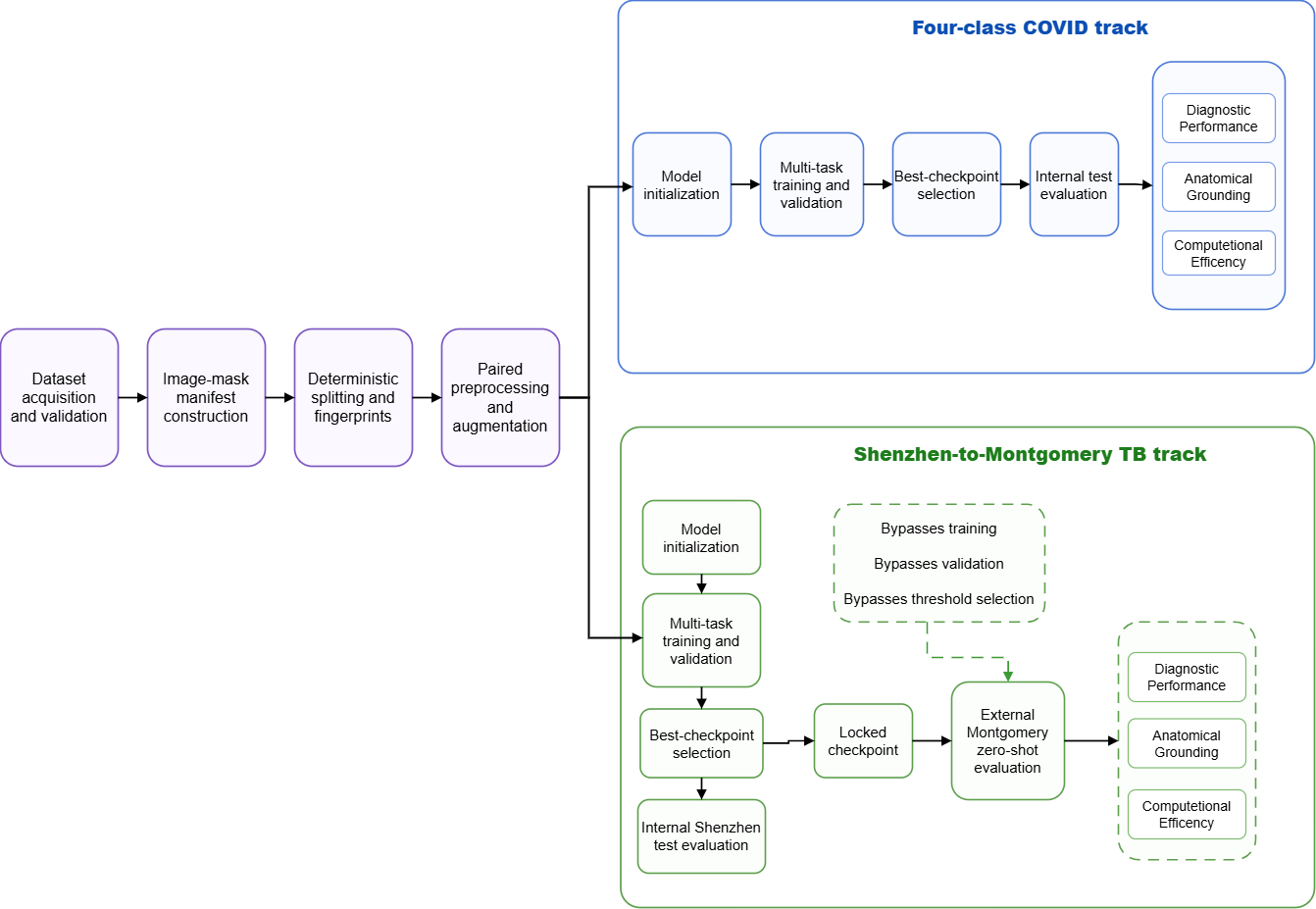}
    \caption{Experimental workflow for four-class COVID-19 evaluation and locked Shenzhen-to-Montgomery TB transfer. The COVID-19 track uses held-out internal testing, whereas the TB track uses Shenzhen for development and checkpoint selection followed by locked Montgomery zero-shot testing. Montgomery is not used for target-domain fine-tuning, recalibration, checkpoint reselection, or threshold optimization.}
    \label{fig:workflow}
\end{figure}

The overall workflow is shown in Fig.~\ref{fig:workflow}. Following data validation, manifest construction, deterministic partitioning, and paired preprocessing, the COVID-19 and TB tracks were executed independently. The TB checkpoint was locked before Montgomery evaluation.

\subsection{Datasets and reference standards}

Three publicly available CXR cohorts were used. Their roles were predefined to separate internal performance assessment from external domain-shift evaluation.

\subsubsection{COVID-19 Radiography Database}

The COVID-19 Radiography Database was assembled through the public releases described by Chowdhury et al.\ \cite{chowdhury2020covid} and Rahman et al.\ \cite{rahman2021enhancement}. The version used here contained 21,165 radiographs: 10,192 Normal, 6,012 Lung Opacity, 3,616 COVID-19, and 1,345 Viral Pneumonia images. A paired lung mask was available for each image. The fixed partition contained 14,815 training images, 3,175 model-selection images, and 3,175 held-out internal-test images. Because reliable patient identifiers are not consistently exposed in the public release, partitioning was performed at image level; this is retained as a study limitation.

\subsubsection{Shenzhen Tuberculosis Dataset}

The Shenzhen dataset contains 662 frontal CXRs, comprising 326 Normal and 336 TB cases \cite{jaeger2014cxr}. The fixed partition contained 463 training images, 99 model-selection images, and 100 held-out internal-test images. Each Shenzhen radiograph was paired with a teacher-generated pulmonary pseudo-mask for anatomy-guided training. These pseudo-masks provide auxiliary anatomical supervision but are not treated as independent expert reference standards.

\subsubsection{Montgomery County Dataset}

The Montgomery County dataset contains 138 frontal CXRs, including 80 Normal and 58 TB cases \cite{jaeger2014cxr}. Separate manual left- and right-lung masks are provided; the pixel-wise union of these masks was used as the pulmonary reference region. All 138 cases were reserved exclusively for external testing.

\begin{table}[!t]
\centering
\caption{Dataset composition and experimental role.}
\label{tab:datasets}
\scriptsize
\setlength{\tabcolsep}{3pt}
\resizebox{\linewidth}{!}{%
\begin{tabular}{l l l r r r r l}
\toprule
Dataset & Task & Classes & Train & Model selection & Internal test & External test & Lung reference \\
\midrule
COVID-19 Radiography & Four-class &
\makecell[l]{Normal, Lung Opacity,\\COVID-19, Viral Pneumonia}
& 14,815 & 3,175 & 3,175 & -- & Paired lung masks \\
Shenzhen & Binary TB & Normal, TB & 463 & 99 & 100 & -- & Teacher pseudo-masks \\
Montgomery & Binary TB & Normal, TB & -- & -- & -- & 138 & Manual left/right lung union \\
\bottomrule
\end{tabular}}
\end{table}

\subsection{Data preparation and reproducibility}

\subsubsection{Manifest construction and integrity checks}

Each cohort was converted into an explicit image--mask manifest containing the radiograph path, diagnostic label, class name, dataset identity, associated mask path or components, and mask provenance. COVID-19 masks were matched using identical filenames, Shenzhen pseudo-masks by filename stem, and Montgomery records required both manual lung-mask components. Missing directories, missing masks, unreadable images, or empty manifests terminated processing rather than silently removing observations.

\subsubsection{Fixed partitioning and multi-seed training}

COVID-19 and Shenzhen were partitioned using a two-stage stratified 70\%/15\%/15\% procedure. The partition was generated once using split seed 42 and remained fixed throughout the multi-seed experiments. For each subset $D_s$, the ordered relative image identifiers were stored and hashed as
\begin{equation}
h_s =
\operatorname{SHA256}
\left(
\operatorname{join}\{r_i:i\in D_s\}
\right),
\end{equation}
where $s\in\{\text{train},\text{selection},\text{test}\}$. The split fingerprints were stored with the experiment metadata and checked on resumed runs. Final training was repeated independently with seeds 12, 42, and 112 while preserving identical data membership.

\subsubsection{Image and mask preprocessing}

Radiographs were loaded in grayscale and converted to three identical channels for ImageNet-pretrained encoders. COVID-19 images underwent contrast-limited adaptive histogram equalization (CLAHE) with clip limit 2.0 and an $8\times8$ tile grid, following the enhancement setting investigated on the same resource \cite{rahman2021enhancement}. Shenzhen images were consumed in their curated stored form, whereas raw Montgomery images received the predefined CLAHE operation before external inference. This dataset-specific rule was fixed before external testing and was not selected using Montgomery labels or outcomes.

Images were resized while preserving aspect ratio so that the longest dimension equaled 224 pixels and were zero-padded to $224\times224$. Lung masks were resized with nearest-neighbor interpolation. ImageNet normalization was applied using
\begin{equation}
\boldsymbol{\mu}=(0.485,0.456,0.406),
\qquad
\boldsymbol{\sigma}=(0.229,0.224,0.225).
\end{equation}
Reference masks were represented as binary $1\times224\times224$ tensors.

Training-only augmentation was applied synchronously to each image--mask pair using Albumentations \cite{buslaev2020albumentations}. The policy included shift and scale changes of up to 5\%, rotation up to $10^\circ$, brightness/contrast perturbation within $\pm0.1$, and horizontal flipping, with probability 0.5 for each augmentation group. Model-selection, internal-test, and external-test images received deterministic preprocessing only.

\subsubsection{Class balancing and deterministic execution}

A weighted random sampler with replacement was used during training. For sample $i$ belonging to class $y_i$,
\begin{equation}
w_i=\frac{1}{n_{y_i}},
\end{equation}
where $n_{y_i}$ denotes the number of training examples in that class. Model-selection and test loaders were not shuffled. Python, NumPy, PyTorch CPU/CUDA generators, data-loader workers, and sampling generators were initialized deterministically for each training seed; cuDNN benchmarking was disabled and deterministic algorithms were requested where supported.

\subsection{Proposed DBCA-SegNet-MGAP framework}

The proposed framework integrates local convolutional representation, hierarchical Transformer context, bidirectional cross-backbone interaction, pulmonary segmentation, and mask-guided classification. ResNet-50 \cite{he2016resnet} was used as the convolutional encoder and Swin-Tiny \cite{liu2021swin} as the hierarchical Transformer encoder. Intermediate representations were exchanged through two BCABs.

\begin{figure}[!t]
    \centering
    \includegraphics[width=\textwidth]{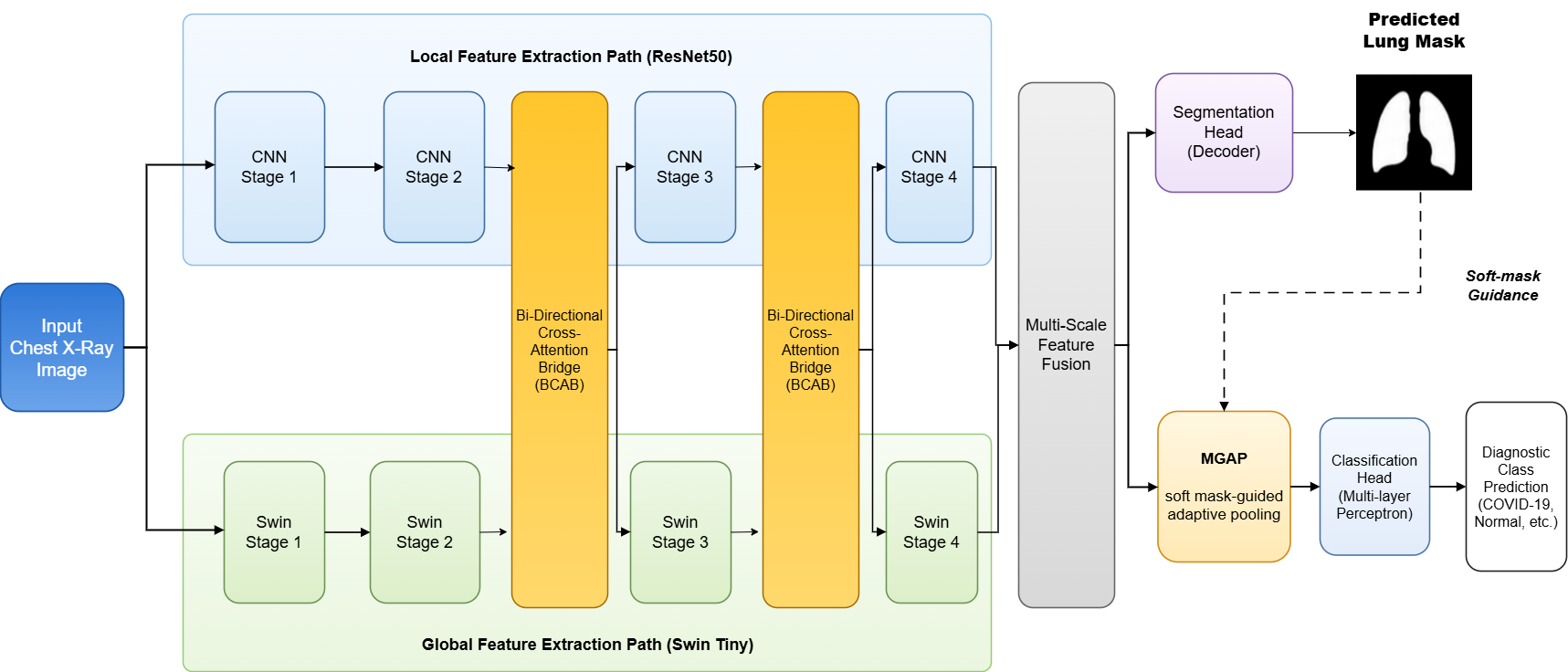}
    \caption{Overall DBCA-SegNet-MGAP architecture. The dual-encoder framework uses BCAB1 at $28\times28$, BCAB2 at $14\times14$, and final $7\times7$ multi-scale feature fusion before pulmonary segmentation and mask-guided classification.}
    \label{fig:architecture}
\end{figure}

\begin{figure}[!t]
    \centering
    \includegraphics[width=\textwidth]{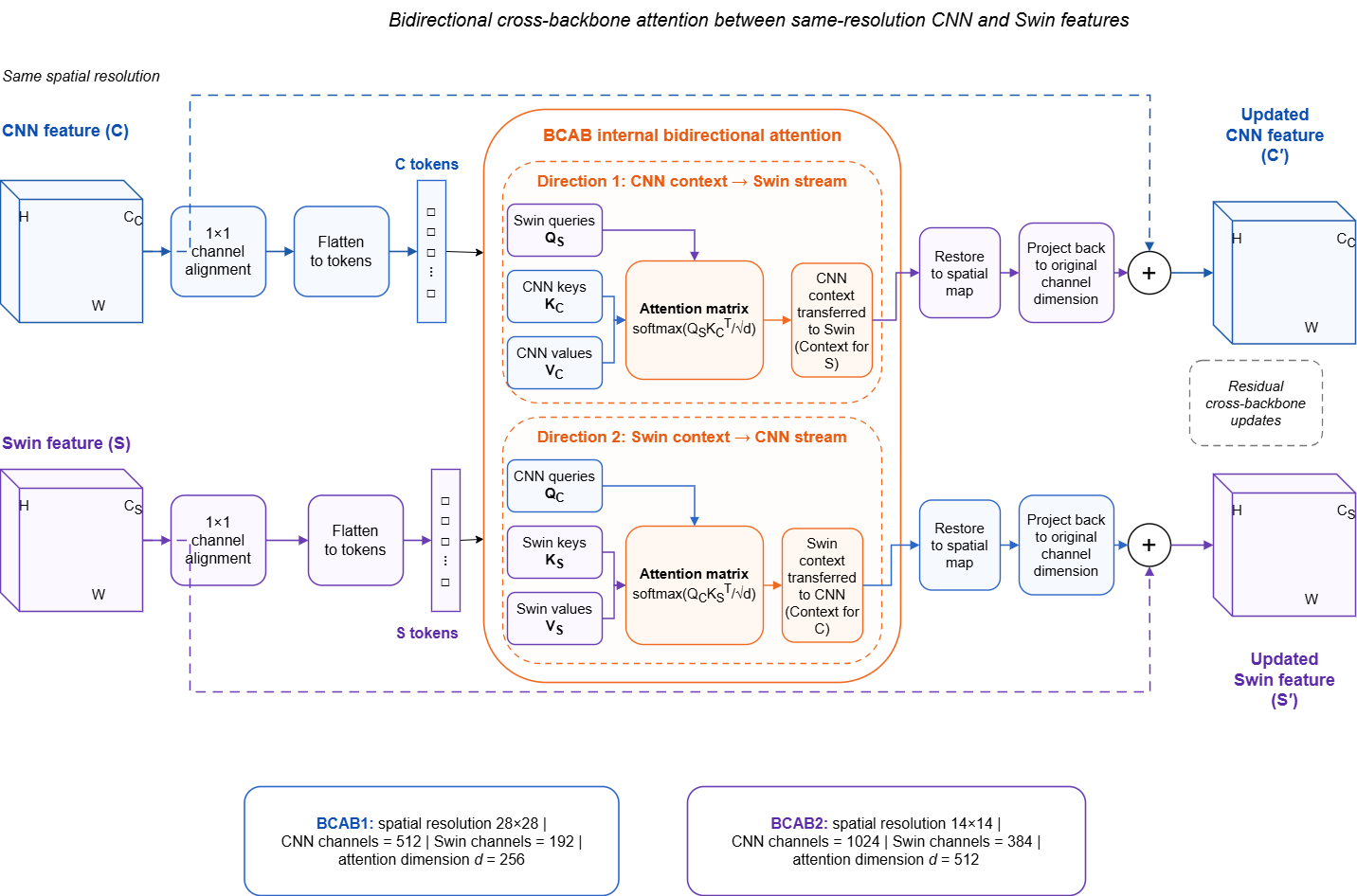}
    \caption{Internal mechanism of the Bidirectional Cross-Backbone Attention Block (BCAB). CNN context is transferred to the Swin stream, and Swin context is transferred to the CNN stream through reciprocal cross-attention and residual feature updates.}
    \label{fig:bcab}
\end{figure}

\begin{figure}[!p]
    \centering
    \includegraphics[width=\textwidth,height=0.78\textheight,keepaspectratio]{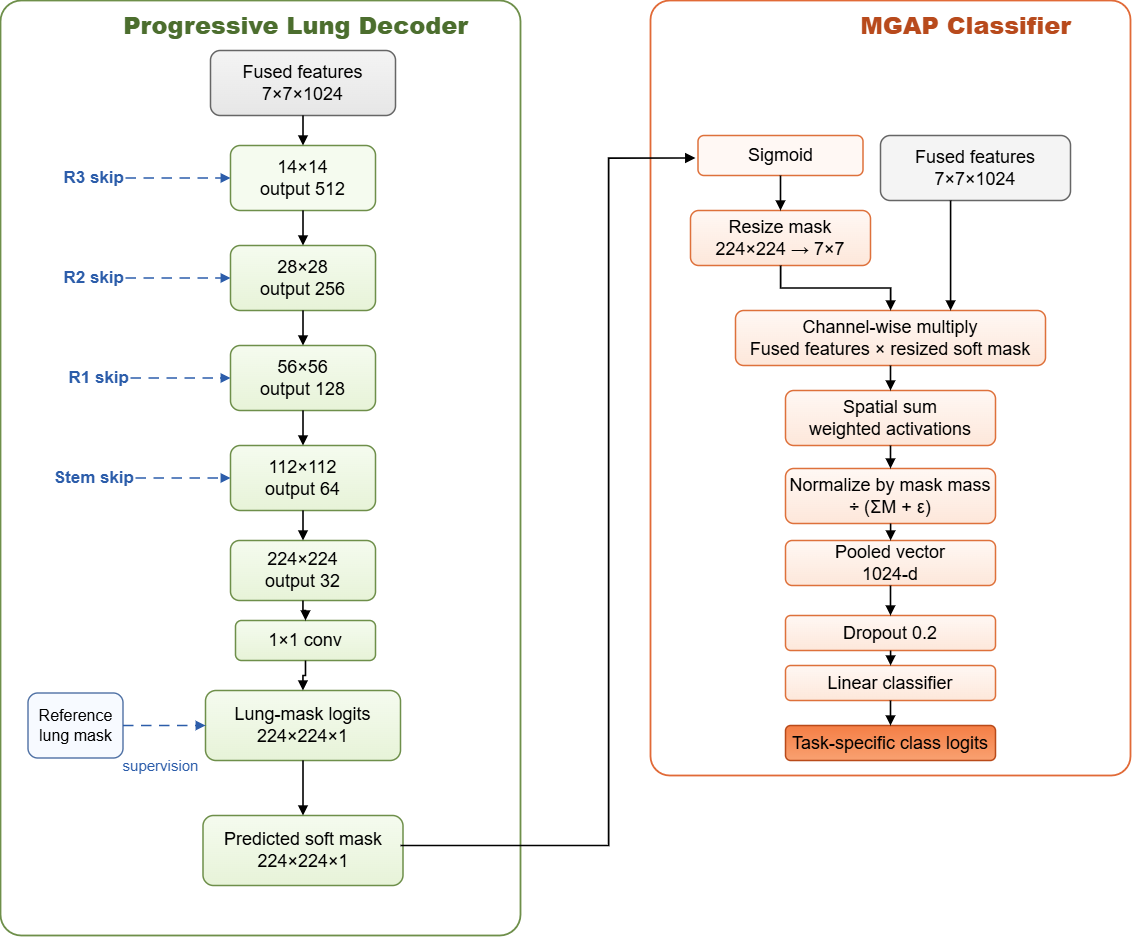}
    \caption{Progressive lung decoder and Mask-Guided Adaptive Global Average Pooling (MGAP) classifier. The decoder reconstructs a soft pulmonary mask from fused features and multiscale ResNet skip connections; MGAP resizes this mask to classifier resolution and uses it to weight normalized spatial feature aggregation.}
    \label{fig:decoder_mgap}
\end{figure}

Figures~\ref{fig:architecture}--\ref{fig:decoder_mgap} separately summarize the overall architecture, BCAB mechanism, and progressive decoder with the MGAP classifier.

\subsubsection{Dual feature encoders}

For an input $x\in\mathbb{R}^{3\times224\times224}$, the ResNet branch produced
\begin{align}
R_1&\in\mathbb{R}^{256\times56\times56},&
R_2&\in\mathbb{R}^{512\times28\times28},\\
R_3&\in\mathbb{R}^{1024\times14\times14},&
R_4&\in\mathbb{R}^{2048\times7\times7},
\end{align}
while the Swin branch produced
\begin{align}
S_1&\in\mathbb{R}^{96\times56\times56},&
S_2&\in\mathbb{R}^{192\times28\times28},\\
S_3&\in\mathbb{R}^{384\times14\times14},&
S_4&\in\mathbb{R}^{768\times7\times7}.
\end{align}
BCAB1 operated on $R_2$ and $S_2$, whereas BCAB2 operated on $R_3$ and $S_3$. The BCAB2-updated ResNet representation was propagated through the final ResNet stage so that the second interaction influenced the final classifier representation.

\subsubsection{Bidirectional Cross-Backbone Attention}

For same-resolution ResNet and Swin features $R$ and $S$, independent $1\times1$ projections aligned channels to an attention width $d$. Flattening the aligned spatial features produced
\begin{equation}
C=\operatorname{tokens}(\phi_R(R)),
\qquad
T=\operatorname{tokens}(\phi_S(S)),
\qquad
C,T\in\mathbb{R}^{N\times d},
\end{equation}
where $N=HW$.

Swin tokens queried CNN keys and values:
\begin{align}
A_{T\leftarrow C}
&=
\operatorname{softmax}
\left(
\frac{Q_T(T)K_C(C)^\top}{\sqrt{d}}
\right),\\
U_T&=A_{T\leftarrow C}V_C(C).
\end{align}
Conversely,
\begin{align}
A_{C\leftarrow T}
&=
\operatorname{softmax}
\left(
\frac{Q_C(C)K_T(T)^\top}{\sqrt{d}}
\right),\\
U_C&=A_{C\leftarrow T}V_T(T).
\end{align}
The context tensors were restored to spatial form, projected back to their original channel widths, and added residually:
\begin{equation}
R'=R+\psi_R(U_C),
\qquad
S'=S+\psi_S(U_T).
\end{equation}
BCAB1 used $d=256$ and BCAB2 used $d=512$.

\subsubsection{Feature fusion and pulmonary decoder}

The final encoder representations were concatenated and projected to a shared feature map:
\begin{equation}
F_0=
\operatorname{ReLU}
\left[
\operatorname{BN}
\left(
\operatorname{Conv}_{1\times1}([R_4;S_4])
\right)
\right],
\qquad
F_0\in\mathbb{R}^{1024\times7\times7}.
\end{equation}
Channel recalibration used a squeeze-and-excitation operation \cite{hu2018senet}:
\begin{equation}
s=
\sigma
\left[
W_2\delta
\left(
W_1\operatorname{GAP}(F_0)
\right)
\right],
\qquad
F=F_0\odot s.
\end{equation}

A progressive decoder reconstructed the lung field from the $7\times7$ fused representation using five upsampling stages. Projected ResNet skip features were incorporated at $14\times14$, $28\times28$, $56\times56$, and $112\times112$, following the multiscale encoder--decoder principle of U-Net \cite{ronneberger2015unet}. A final $1\times1$ convolution produced
\begin{equation}
G\in\mathbb{R}^{1\times224\times224}
\end{equation}
lung-mask logits. The continuous probability map $\sigma(G)$ was retained for classification rather than converted into a hard crop.

\subsubsection{Mask-Guided Adaptive Global Average Pooling}

Let $F\in\mathbb{R}^{C\times h\times w}$ denote the final classifier feature representation and
\begin{equation}
\widetilde{M}
=
\operatorname{Resize}_{h,w}\!\left[\sigma(G)\right],
\qquad h=w=7,
\end{equation}
the predicted soft lung mask at classifier resolution. MGAP computes
\begin{equation}
z_c=
\frac{
\displaystyle\sum_{i=1}^{h}\sum_{j=1}^{w}
F_{cij}\widetilde{M}_{ij}
}{
\displaystyle\sum_{i=1}^{h}\sum_{j=1}^{w}
\widetilde{M}_{ij}+\epsilon
},
\qquad
\epsilon=10^{-6}.
\end{equation}
The mask therefore acts as a continuous spatial weighting mechanism rather than a hard crop. The pooled vector is passed through dropout ($p=0.2$) and a linear classifier:
\begin{equation}
\hat{\mathbf{y}}
=
W_{\mathrm{cls}}\operatorname{Dropout}(z)
+b_{\mathrm{cls}}.
\end{equation}
The output dimension is four for the COVID-19 task and two for TB. Architecture-matched GAP controls omit the soft-mask weighting while retaining the same preceding representation.

\subsection{Multi-task training and optimization}

Disease classification was optimized using cross-entropy with label smoothing $\alpha=0.05$. For reference class $y$,
\begin{equation}
q_k=(1-\alpha)\mathbb{I}[k=y]+\frac{\alpha}{K},
\end{equation}
and
\begin{equation}
\mathcal{L}_{\mathrm{cls}}
=
-\frac{1}{B}
\sum_{b=1}^{B}\sum_{k=1}^{K}
q_{bk}\log p_{bk}.
\end{equation}

The pulmonary decoder was optimized using binary cross-entropy with logits plus soft Dice loss. For $\hat m_i=\sigma(g_i)$ and reference pixel $m_i$,
\begin{equation}
\operatorname{Dice}_{\mathrm{soft}}
=
\frac{
2\sum_i \hat m_i m_i+\epsilon
}{
\sum_i \hat m_i+\sum_i m_i+\epsilon
},
\end{equation}
\begin{equation}
\mathcal{L}_{\mathrm{seg}}
=
\mathcal{L}_{\mathrm{BCE}}
+
\left(
1-\operatorname{Dice}_{\mathrm{soft}}
\right).
\end{equation}
The joint objective was
\begin{equation}
\mathcal{L}_{\mathrm{total}}
=
\mathcal{L}_{\mathrm{cls}}
+
\lambda_{\mathrm{seg}}\mathcal{L}_{\mathrm{seg}},
\qquad
\lambda_{\mathrm{seg}}=0.2.
\end{equation}

Models were trained for 30 epochs using AdamW \cite{loshchilov2019adamw}. The initial learning rate was $2\times10^{-4}$ and was cosine-annealed toward $10^{-6}$. A physical batch size of 8 with two-step gradient accumulation produced an effective batch size of 16. Gradients were clipped to a maximum norm of 1.0, training used FP32 precision, and the best checkpoint was selected by model-selection weighted F1. Experiments were implemented in PyTorch and executed on an NVIDIA Tesla T4 GPU.

\begin{table}[!t]
\centering
\caption{Shared training configuration.}
\label{tab:training}
\small
\begin{tabular}{ll}
\toprule
Setting & Value \\
\midrule
Input & $224\times224\times3$ \\
Initialization & ImageNet-pretrained encoders \\
Epochs & 30 \\
Optimizer & AdamW \\
Learning rate & $2\times10^{-4}\rightarrow10^{-6}$ \\
Weight decay & $1\times10^{-4}$ \\
Batch size & 8 physical / 16 effective \\
Loss & $\mathcal{L}_{cls}+0.2\mathcal{L}_{seg}$ \\
Label smoothing & 0.05 \\
Dropout & 0.20 \\
Gradient clipping & 1.0 \\
Checkpoint criterion & Model-selection weighted F1 \\
Data-partition seed & 42 \\
Training seeds & 12, 42, 112 \\
\bottomrule
\end{tabular}
\end{table}

\subsection{Baselines and controlled experimental design}

Four pretrained classifiers were retained as conventional diagnostic baselines: ResNet-50 \cite{he2016resnet}, Swin-Tiny \cite{liu2021swin}, DenseNet-121 \cite{huang2017densenet}, and ConvNeXt-Tiny \cite{liu2022convnext}. Each used its pretrained feature extractor, conventional GAP, dropout, and a task-specific linear classifier.

The mechanism experiment used a factorial architecture-matched design. Three full-capacity representation conditions were defined: late fusion, single-bridge BCAB1, and dual-bridge BCAB1+BCAB2. Each representation was paired with either GAP or MGAP. The lightweight capacity-control pair used EfficientNet-B0 \cite{tan2019efficientnet} and TinyViT-5M \cite{wu2022tinyvit} with final-stage fusion and no BCAB. It was not used to infer the BCAB effect.

\begin{table}[!t]
\centering
\caption{Controlled experimental configurations.}
\label{tab:controls}
\small
\setlength{\tabcolsep}{4pt}
\resizebox{\linewidth}{!}{%
\begin{tabular}{l l c c c l}
\toprule
Representation & Encoders & BCAB1 & BCAB2 & Pooling & Experimental role \\
\midrule
Late fusion & ResNet-50 + Swin-Tiny & -- & -- & GAP / MGAP & No-bridge mechanism control \\
Single bridge & ResNet-50 + Swin-Tiny & \cmark & -- & GAP / MGAP & Intermediate-fusion control \\
Dual bridge & ResNet-50 + Swin-Tiny & \cmark & \cmark & GAP / MGAP & Complete architecture \\
Lightweight & EfficientNet-B0 + TinyViT-5M & -- & -- & GAP / MGAP & Capacity control \\
\bottomrule
\end{tabular}}
\end{table}

\subsection{Evaluation and statistical analysis}

\subsubsection{Diagnostic performance and calibration}

Primary diagnostic metrics were accuracy, weighted F1, macro F1, and ROC-AUC. Weighted F1 accounts for the imbalanced COVID-19 class distribution, while macro F1 assigns equal contribution to each class. For binary TB testing, sensitivity and specificity at the fixed threshold 0.5 were treated as secondary threshold-dependent measures. Multiple metrics were retained because metric choice can change the interpretation and ranking of medical-imaging systems \cite{reinke2024metrics}.

Calibration was summarized using 15-bin Expected Calibration Error (ECE):
\begin{equation}
\operatorname{ECE}
=
\sum_{m=1}^{15}
\frac{|B_m|}{N}
\left|
\operatorname{acc}(B_m)-\operatorname{conf}(B_m)
\right|,
\end{equation}
where $B_m$ is the set of samples assigned to confidence bin $m$.

\subsubsection{Pulmonary segmentation}

Predicted lung masks were thresholded at 0.5 for segmentation evaluation. The primary segmentation metric was Dice:
\begin{equation}
\operatorname{Dice}(P,M)
=
\frac{2|P\cap M|+\epsilon}{|P|+|M|+\epsilon}.
\end{equation}
COVID-19 predictions were evaluated against paired dataset masks, Shenzhen predictions against teacher-generated pseudo-masks, and Montgomery predictions against the union of manual left- and right-lung masks. These measurements evaluate lung-field segmentation rather than disease-lesion localization.

\subsubsection{Pulmonary attribution containment}

Grad-CAM \cite{selvaraju2017gradcam} was used for the primary quantitative containment analysis. In DBCA-SegNet-MGAP, the Grad-CAM target was the $7\times7$ mask-influenced classifier representation immediately before MGAP; the matched GAP control used the corresponding unweighted fused representation. Layer-CAM \cite{jiang2021layercam} was generated from the BCAB2-updated $14\times14$ ResNet representation as a complementary higher-resolution qualitative view. Attribution maps were upsampled to image resolution and min--max normalized.

Let $A(i)\geq0$ denote normalized predicted-class Grad-CAM attribution and $M(i)\in\{0,1\}$ the pulmonary reference mask. ALR was defined as
\begin{equation}
\operatorname{ALR}
=
\frac{\sum_i A(i)M(i)}
{\sum_i A(i)+\epsilon}.
\end{equation}
High-intensity containment was summarized using
\begin{equation}
\operatorname{cALR}@\tau
=
\frac{
\sum_i A(i)\mathbb{I}[A(i)\geq\tau]M(i)
}{
\sum_i A(i)\mathbb{I}[A(i)\geq\tau]+\epsilon
}.
\end{equation}
The main analysis reports cALR@0.9. It describes where attribution mass surviving the 0.9 threshold is located; it does not represent the fraction of the lung attended to and is not interpreted as lesion localization, explanation faithfulness, or causal evidence.

For selected qualitative TB cases, normalized class-specific Grad-CAM maps were probability weighted:
\begin{equation}
A_y^{w}(x)=P(y\mid x)A_y(x).
\end{equation}
The signed difference
\begin{equation}
D(x)=A_{\mathrm{TB}}^{w}(x)-A_{\mathrm{Normal}}^{w}(x)
\end{equation}
was visualized to examine spatial class dominance. This analysis was descriptive and was not used as a quantitative endpoint.

\subsubsection{Multi-seed variability}

Principal experiments were independently repeated with training seeds 12, 42, and 112 on the same fixed data partition. Results are summarized as mean $\pm$ sample SD. The across-seed SD characterizes stochastic training variability; with only three runs, it is not interpreted as a formal confidence interval or evidence of statistical superiority.

\subsubsection{Locked external zero-shot evaluation}

For the external experiment, model development and checkpoint selection used Shenzhen only. The checkpoint with the highest Shenzhen model-selection weighted F1 was frozen before Montgomery evaluation. The same checkpoint was applied to all 138 Montgomery radiographs with no target-domain training, no fine-tuning, no target-specific checkpoint selection, no probability recalibration, no threshold optimization, and a fixed TB decision threshold of 0.5. ROC-AUC characterized external discrimination, weighted F1 and threshold-dependent metrics characterized operating-point transfer, ECE characterized calibration shift, and pulmonary attribution was reassessed using the same ALR/cALR definitions.

\subsubsection{Model complexity and inference efficiency}

Model complexity was summarized using trainable parameter count, serialized checkpoint size, and inference throughput:
\begin{equation}
\operatorname{Throughput}
=
\frac{N_{\mathrm{images}}}{t_{\mathrm{inference}}},
\end{equation}
after GPU synchronization. Throughput is hardware- and implementation-dependent and is therefore interpreted only as a within-study comparison.

% ============================================================
\section{Results}
\label{sec:results}
% ============================================================

All principal internal model comparisons and COVID-19 mechanism experiments were evaluated over three training runs using seeds 12, 42, and 112 while retaining the fixed partitions described in Section~\ref{sec:methods}. Unless otherwise stated, results are reported as mean $\pm$ sample SD. The SD characterizes run-to-run variability and is not interpreted as a formal test of statistical superiority.

\subsection{Internal diagnostic performance and pulmonary attribution}

DBCA-SegNet-MGAP showed stable diagnostic performance on the four-class COVID-19 test set. Across the three runs, accuracy was $0.9616\pm0.0016$, weighted F1 was $0.9615\pm0.0015$, macro F1 was $0.9686\pm0.0014$, macro ROC-AUC was $0.9906\pm0.0007$, and ECE was $0.0247\pm0.0013$.

\begin{figure}[!t]
    \centering
    \includegraphics[width=\textwidth]{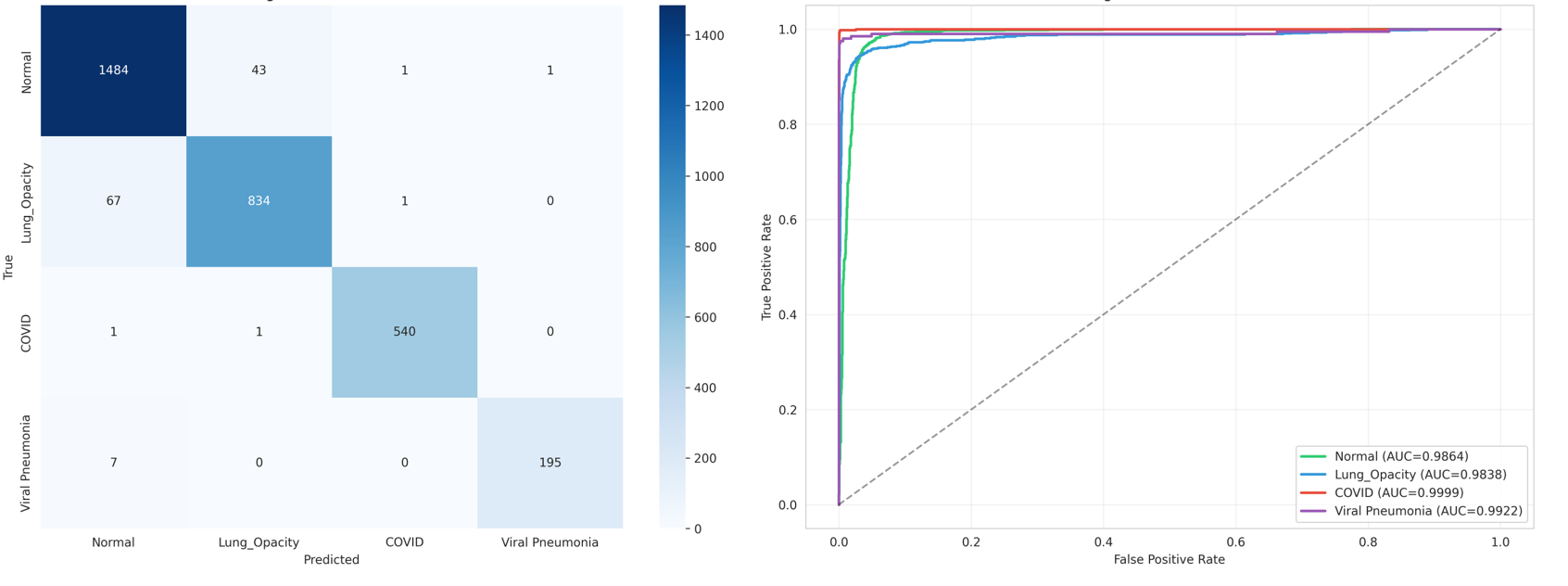}
    \caption{Representative Seed-42 diagnostic behavior of DBCA-SegNet-MGAP on the held-out four-class COVID-19 test set: confusion matrix (left) and one-vs-rest ROC curves (right). The figure is shown for qualitative error-pattern inspection; multi-seed aggregate metrics are reported in Table~\ref{tab:internal}.}
    \label{fig:covid_diagnostics}
\end{figure}

At class level, COVID-19 was the most consistently classified category, with F1 $=0.9963\pm0.0009$. Viral Pneumonia achieved $0.9798\pm0.0011$, Normal $0.9611\pm0.0013$, and Lung Opacity $0.9372\pm0.0024$. Lung Opacity therefore remained the most difficult class. Figure~\ref{fig:covid_diagnostics} shows the representative Seed-42 confusion matrix and one-vs-rest ROC curves; aggregate multi-seed results are reported separately in Table~\ref{tab:internal}.

\begin{table}[!t]
\centering
\caption{Multi-seed COVID-19 diagnostic performance, calibration,
and pulmonary-attribution comparison.}
\label{tab:internal}
\small
\setlength{\tabcolsep}{4pt}

\resizebox{\linewidth}{!}{%
\begin{tabular}{lcccccc}
\toprule
\multirow{2}{*}{Configuration}
& \multicolumn{4}{c}{Diagnostic performance and calibration}
& \multicolumn{2}{c}{Pulmonary attribution} \\
\cmidrule(lr){2-5}
\cmidrule(lr){6-7}
& Acc.
& F1$_w$
& AUC
& ECE
& ALR
& cALR@0.9 \\
\midrule

ResNet-50
& $0.9545\pm0.0017$ & $0.9546\pm0.0017$ & $0.9925\pm0.0007$ & $0.0298\pm0.0013$ & $0.4981\pm0.0100$ & $0.7659\pm0.0107$ \\

Swin-Tiny
& $0.9628\pm0.0014$ & $0.9627\pm0.0014$ & $\mathbf{0.9934\pm0.0007}$ & $0.0307\pm0.0013$ & $0.2415\pm0.0061$ & $0.2054\pm0.0065$ \\

DenseNet-121
& $0.9620\pm0.0011$ & $0.9620\pm0.0011$ & $0.9911\pm0.0007$ & $0.0308\pm0.0014$ & $0.3228\pm0.0078$ & $0.3546\pm0.0083$ \\

ConvNeXt-Tiny
& $0.9560\pm0.0012$ & $0.9560\pm0.0012$ & $0.9915\pm0.0008$ & $0.0315\pm0.0013$ & $0.2679\pm0.0074$ & $0.2342\pm0.0072$ \\

DBCA-Light + GAP
& $0.9615\pm0.0011$ & $0.9614\pm0.0011$ & $0.9911\pm0.0006$ & $\mathbf{0.0247\pm0.0015}$ & $0.2758\pm0.0077$ & $0.3758\pm0.0091$ \\

DBCA-Light + MGAP
& $\mathbf{0.9664\pm0.0011}$ & $\mathbf{0.9664\pm0.0011}$ & $0.9899\pm0.0008$ & $0.0254\pm0.0015$ & $0.4482\pm0.0101$ & $0.8572\pm0.0097$ \\

\textbf{DBCA-SegNet-MGAP}
& $0.9616\pm0.0016$ & $0.9615\pm0.0015$ & $0.9906\pm0.0007$ & $0.0247\pm0.0013$ & $\mathbf{0.7086\pm0.0104}$ & $\mathbf{0.9905\pm0.0018}$ \\

\bottomrule
\end{tabular}}

\vspace{0.3em}
\footnotesize
Values are mean $\pm$ sample SD across seeds 12, 42, and 112.
Bold values denote the highest displayed mean point estimate and do not
imply statistical significance.
\end{table}

The diagnostic and pulmonary-attribution metrics produced different model rankings. DBCA-Light + MGAP achieved the highest mean accuracy and weighted F1, while Swin-Tiny achieved the highest macro AUC. DBCA-SegNet-MGAP was therefore not the highest-performing classifier on every conventional endpoint.

In contrast, the complete configuration produced the strongest pulmonary containment among the displayed configurations. Its ALR of $0.7086\pm0.0104$ exceeded ResNet-50 ($0.4981\pm0.0100$), DBCA-Light + MGAP ($0.4482\pm0.0101$), and the remaining baselines. The contrast with Swin-Tiny is particularly informative: Swin-Tiny achieved slightly higher weighted F1 ($0.9627$ vs.\ $0.9615$) and macro AUC ($0.9934$ vs.\ $0.9906$), but ALR was $0.2415$ rather than $0.7086$. Thus, within the evaluated protocol, high diagnostic competence did not uniquely determine pulmonary-attribution behavior.

\subsection{Controlled effect of mask-guided pooling}

To isolate mask-guided aggregation, GAP and MGAP were compared within late-fusion, single-bridge, and dual-bridge representation conditions. Table~\ref{tab:mgap_effect} reports the direct GAP$\rightarrow$MGAP effect within each matched representation.

\begin{table}[!t]
\centering
\caption{Architecture-matched effect of replacing GAP with MGAP on the COVID-19 test set.}
\label{tab:mgap_effect}
\small
\setlength{\tabcolsep}{4pt}
\resizebox{\linewidth}{!}{%
\begin{tabular}{l l c c c}
\toprule
Representation & Metric & GAP & MGAP & $\Delta$ (MGAP $-$ GAP) \\
\midrule
\multirow{4}{*}{Late fusion}
& Weighted F1 & $0.9579\pm0.0012$ & $0.9609\pm0.0014$ & $+0.0029$ \\
& Dice & $0.9780\pm0.0011$ & $0.9764\pm0.0012$ & $-0.0016$ \\
& ALR & $0.2668\pm0.0074$ & $\mathbf{0.5884\pm0.0118}$ & $\mathbf{+0.3216}$ \\
& cALR@0.9 & $0.3685\pm0.0122$ & $\mathbf{0.9116\pm0.0079}$ & $\mathbf{+0.5431}$ \\
\midrule
\multirow{4}{*}{Single bridge (BCAB1)}
& Weighted F1 & $0.9627\pm0.0014$ & $0.9650\pm0.0012$ & $+0.0023$ \\
& Dice & $0.9846\pm0.0010$ & $0.9793\pm0.0012$ & $-0.0053$ \\
& ALR & $0.2940\pm0.0076$ & $\mathbf{0.6268\pm0.0121}$ & $\mathbf{+0.3328}$ \\
& cALR@0.9 & $0.5244\pm0.0104$ & $\mathbf{0.9587\pm0.0059}$ & $\mathbf{+0.4343}$ \\
\midrule
\multirow{4}{*}{Dual bridge (BCAB1 + BCAB2)}
& Weighted F1 & $0.9618\pm0.0015$ & $0.9615\pm0.0015$ & $-0.0003$ \\
& Dice & $0.9849\pm0.0010$ & $0.9812\pm0.0014$ & $-0.0037$ \\
& ALR & $0.3878\pm0.0098$ & $\mathbf{0.7086\pm0.0104}$ & $\mathbf{+0.3209}$ \\
& cALR@0.9 & $0.5265\pm0.0101$ & $\mathbf{0.9905\pm0.0018}$ & $\mathbf{+0.4640}$ \\
\bottomrule
\end{tabular}}
\end{table}

The anatomical effect of MGAP was directionally consistent across all three representation conditions. Mean ALR increased by $+0.3216$ under late fusion, $+0.3328$ with BCAB1, and $+0.3209$ with the complete dual-bridge representation. cALR@0.9 increased by $+0.5431$, $+0.4343$, and $+0.4640$, respectively. By comparison, weighted F1 changed by only $+0.0029$, $+0.0023$, and $-0.0003$.

The dual-bridge comparison provides the clearest controlled result. Replacing GAP with MGAP increased ALR from $0.3878\pm0.0098$ to $0.7086\pm0.0104$ and cALR@0.9 from $0.5265\pm0.0101$ to $0.9905\pm0.0018$, while weighted F1 remained essentially unchanged. The ALR increases were $+0.3197$, $+0.3219$, and $+0.3210$ for seeds 12, 42, and 112, respectively.

Segmentation and classifier attribution were also not equivalent. In the dual-bridge condition, GAP achieved slightly higher Dice than MGAP ($0.9849\pm0.0010$ vs.\ $0.9812\pm0.0014$) while producing substantially lower ALR ($0.3878$ vs.\ $0.7086$). High-quality lung-mask prediction therefore did not, by itself, imply that class-related attribution was concentrated within the lungs.

\begin{figure}[!t]
    \centering
    \includegraphics[width=0.90\textwidth]{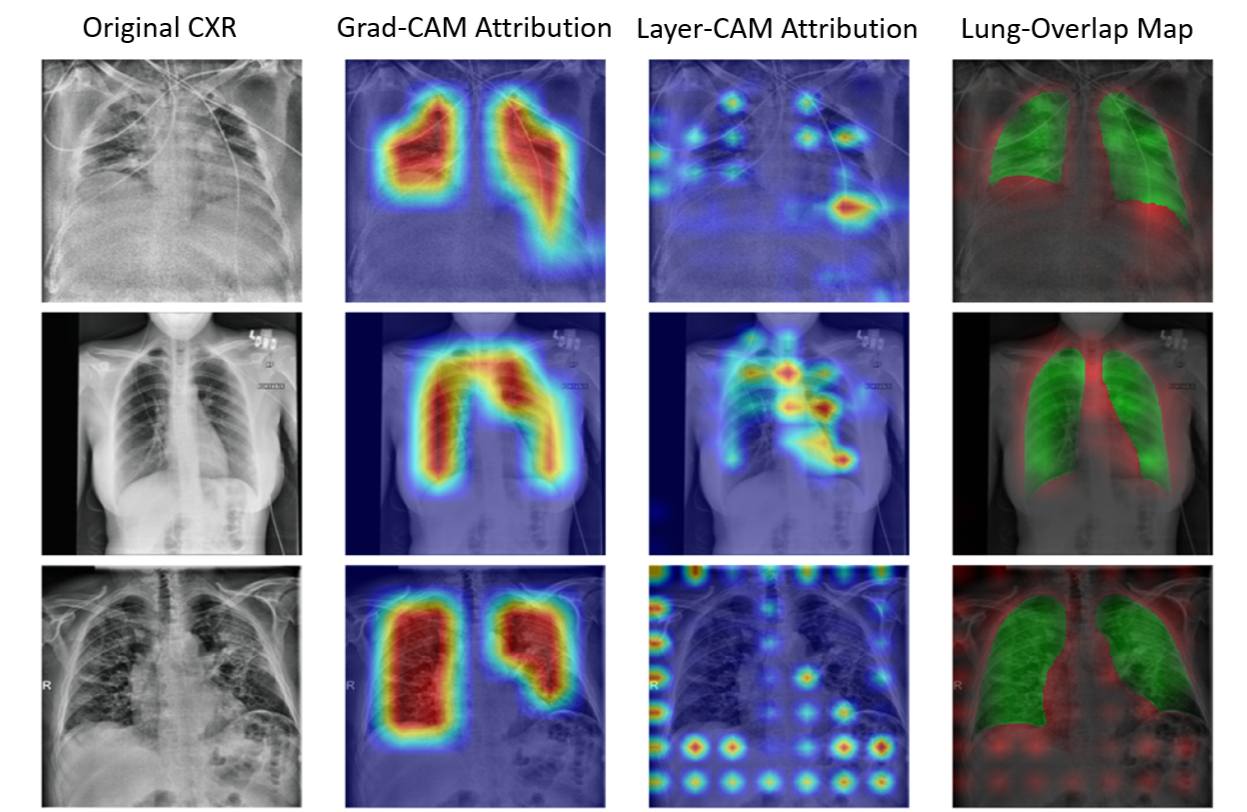}
    \caption{Representative pulmonary-attribution analysis of DBCA-SegNet-MGAP using Grad-CAM, Layer-CAM, and lung-overlap maps. Columns show the original CXR, predicted-class Grad-CAM attribution, Layer-CAM attribution, and the attribution--lung overlap map. Green indicates attribution within the pulmonary reference region and red indicates extrapulmonary attribution.}
    \label{fig:pulm_attr}
\end{figure}

Representative attribution maps are shown in Fig.~\ref{fig:pulm_attr}. Grad-CAM provides broad predicted-class attribution, Layer-CAM provides a finer intermediate-layer view, and the overlap map separates pulmonary-contained from extrapulmonary attribution. These images complement the cohort-level ALR/cALR analysis but are not treated as lesion annotations or causal explanations.

\subsection{External Shenzhen-to-Montgomery zero-shot evaluation}

External reliability was evaluated by training and selecting models exclusively on Shenzhen and applying the locked checkpoints to Montgomery without target-domain fine-tuning, recalibration, or threshold optimization.

For DBCA-SegNet-MGAP, Shenzhen weighted F1 was $0.8898\pm0.0034$ and ROC-AUC was $0.9684\pm0.0019$. Under zero-shot transfer to Montgomery, weighted F1 decreased to $0.7528\pm0.0080$, whereas ROC-AUC remained $0.9080\pm0.0043$. ECE increased from $0.0715\pm0.0034$ to $0.1683\pm0.0055$. The mean source-to-target changes were therefore $-0.1371$ in weighted F1, $-0.0604$ in ROC-AUC, and $+0.0968$ in ECE. Montgomery weighted F1 ranged from 0.7449 to 0.7608 across the three runs, ROC-AUC from 0.9037 to 0.9123, and ECE from 0.1627 to 0.1736.

\begin{table}[!t]
\centering
\caption{Multi-seed Shenzhen internal and Montgomery zero-shot external
performance and pulmonary-attribution comparison.}
\label{tab:external}
\small
\setlength{\tabcolsep}{5pt}

\resizebox{\linewidth}{!}{%
\begin{tabular}{llcccc}
\toprule
Configuration
& Cohort
& F1$_w$
& AUC
& ALR
& cALR@0.9 \\
\midrule

\multirow{2}{*}{ResNet-50}
& Shenzhen
& $0.7885\pm0.0033$
& $0.8657\pm0.0032$
& --
& -- \\
& Montgomery
& $0.4496\pm0.0068$
& $0.5698\pm0.0082$
& $0.4290\pm0.0074$
& $0.3261\pm0.0114$ \\
\midrule

\multirow{2}{*}{Swin-Tiny}
& Shenzhen
& $0.8493\pm0.0033$
& $0.9259\pm0.0032$
& --
& -- \\
& Montgomery
& $0.7345\pm0.0067$
& $0.8532\pm0.0062$
& $0.2498\pm0.0068$
& $0.1868\pm0.0077$ \\
\midrule

\multirow{2}{*}{DenseNet-121}
& Shenzhen
& $0.8797\pm0.0032$
& $0.9206\pm0.0028$
& --
& -- \\
& Montgomery
& $0.7299\pm0.0063$
& $0.7707\pm0.0066$
& $0.4379\pm0.0073$
& $0.4244\pm0.0083$ \\
\midrule

\multirow{2}{*}{DBCA-Light + GAP}
& Shenzhen
& $0.8490\pm0.0032$
& $0.9300\pm0.0030$
& --
& -- \\
& Montgomery
& $0.7509\pm0.0068$
& $0.8601\pm0.0059$
& $0.3468\pm0.0072$
& $0.5370\pm0.0082$ \\
\midrule

\multirow{2}{*}{DBCA-Light + MGAP}
& Shenzhen
& $0.8791\pm0.0033$
& $0.9360\pm0.0033$
& --
& -- \\
& Montgomery
& $0.5440\pm0.0069$
& $0.8352\pm0.0059$
& $0.6293\pm0.0079$
& $0.9853\pm0.0032$ \\
\midrule

\multirow{2}{*}{\textbf{DBCA-SegNet-MGAP}}
& Shenzhen
& $\mathbf{0.8898\pm0.0034}$
& $\mathbf{0.9684\pm0.0019}$
& --
& -- \\
& Montgomery
& $\mathbf{0.7528\pm0.0080}$
& $\mathbf{0.9080\pm0.0043}$
& $\mathbf{0.6466\pm0.0081}$
& $\mathbf{0.9905\pm0.0020}$ \\

\bottomrule
\end{tabular}}

\vspace{0.3em}
\footnotesize
Values are mean $\pm$ sample SD across seeds 12, 42, and 112.
Pulmonary-attribution metrics are shown for the external Montgomery cohort,
which is the principal domain-shift attribution analysis.
\end{table}

The complete framework achieved the highest mean point estimates for source-domain F1 and AUC and for external AUC among the displayed configurations. On Montgomery, weighted F1 ($0.7528\pm0.0080$) was close to DBCA-Light + GAP ($0.7509\pm0.0068$), while ROC-AUC was higher ($0.9080\pm0.0043$ vs.\ $0.8601\pm0.0059$). Pulmonary attribution containment remained high, with ALR $=0.6466\pm0.0081$ and cALR@0.9 $=0.9905\pm0.0020$.

The capacity-control pair provides an important counterexample. Replacing GAP with MGAP in DBCA-Light increased Montgomery ALR from $0.3468\pm0.0072$ to $0.6293\pm0.0079$ and cALR@0.9 from $0.5370\pm0.0082$ to $0.9853\pm0.0032$, yet weighted F1 decreased from $0.7509\pm0.0068$ to $0.5440\pm0.0069$, and AUC decreased from $0.8601\pm0.0059$ to $0.8352\pm0.0059$. Strong pulmonary containment therefore did not guarantee robust external classification.

\begin{figure}[!t]
    \centering
    \includegraphics[width=\textwidth]{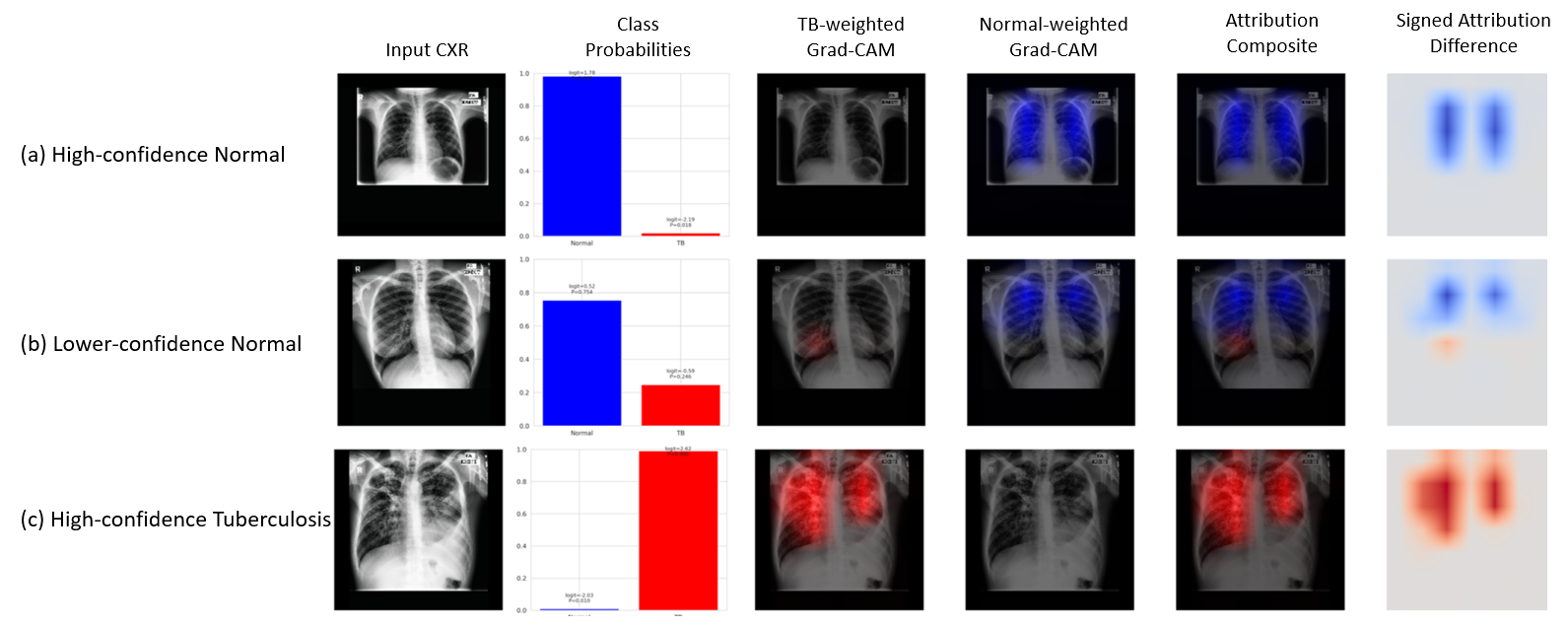}
    \caption{Representative probability-weighted competing-class attribution for Tuberculosis versus Normal under external Montgomery testing. Rows show a high-confidence Normal prediction, a lower-confidence Normal prediction, and a high-confidence Tuberculosis prediction. Columns show the input CXR, class probabilities, TB-weighted Grad-CAM, Normal-weighted Grad-CAM, attribution composite, and signed attribution difference.}
    \label{fig:tb_competing}
\end{figure}

Figure~\ref{fig:tb_competing} provides a qualitative view of probability-weighted competing-class attribution on Montgomery. Red corresponds to TB-dominant attribution and blue to Normal-dominant attribution in the signed-difference map. These maps are descriptive and are not interpreted as verified lesion localization or causal evidence.

\subsection{Model complexity and inference efficiency}

Model complexity and inference efficiency were evaluated to contextualize the capacity controls.

\begin{table}[!t]
\centering
\caption{Model complexity and inference efficiency of the capacity-controlled DBCA configurations.}
\label{tab:efficiency}
\small
\setlength{\tabcolsep}{4pt}
\resizebox{\linewidth}{!}{%
\begin{tabular}{l c c c}
\toprule
Configuration & Parameters (M) & Checkpoint (MB) & Throughput (img/s) \\
\midrule
DBCA-Light + GAP & $\mathbf{9.028}$ & $\mathbf{34.88}$ & $\mathbf{158.83}$ \\
DBCA-Light + MGAP & $\mathbf{9.028}$ & $\mathbf{34.88}$ & $89.98$ \\
DBCA-SegNet + GAP & $68.821$ & $262.96$ & $58.60$ \\
\textbf{DBCA-SegNet + MGAP} & $68.821$ & $262.96$ & $59.33$ \\
\bottomrule
\end{tabular}}

\vspace{0.3em}
\footnotesize Throughput was measured using the common COVID-19 inference benchmark and is intended for within-study comparison.
\end{table}

The lightweight configurations contained 9.028 million parameters, an approximately 86.9\% reduction relative to the 68.821-million-parameter complete architecture, while checkpoint size decreased from 262.96 MB to 34.88 MB. Within the complete architecture, replacing GAP with MGAP did not increase parameter count or checkpoint size, and measured throughput remained similar (58.60 vs.\ 59.33 images/s). Thus, the increase in pulmonary attribution containment associated with MGAP was not attributable to increased parameter count.

The lightweight configurations show that substantially lower complexity is achievable, but Sections~\ref{sec:results} and Table~\ref{tab:external} show that reduced capacity did not reproduce the complete framework's overall external reliability profile. Efficiency and reliability were therefore treated as separate properties.

% ============================================================
\section{Discussion}
\label{sec:discussion}
% ============================================================

\subsection{Diagnostic competence and pulmonary attribution are separable}

The internal experiments directly answer RQ1: classifiers with similar diagnostic performance can exhibit substantially different pulmonary-attribution distributions. Swin-Tiny slightly exceeded DBCA-SegNet-MGAP in weighted F1 and macro AUC, yet its mean ALR was less than half that of the complete framework. Conversely, DBCA-Light + MGAP achieved the highest internal weighted F1 while retaining substantially lower ALR than DBCA-SegNet-MGAP. These differences support the central premise that conventional diagnostic metrics do not determine where class-related attribution is spatially concentrated.

This result is consistent with the broader caution that high CXR performance can coexist with unintended shortcut behavior \cite{degrave2021shortcuts}. It also complements anatomy-guided classification studies \cite{zhang2021pmgan,agfrnet,wuAnatomyGuided}: the relevant question is not only whether anatomical information can be introduced, but whether its influence on the classifier can be isolated and measured. ALR and cALR provide one reproducible containment-oriented view of that question, while remaining deliberately narrower than lesion localization or explanation faithfulness.

\subsection{MGAP as a controlled anatomical aggregation intervention}

The matched GAP/MGAP experiments address RQ2. Across late-fusion, single-bridge, and dual-bridge representations, MGAP increased ALR by approximately 0.32 in all three conditions and produced large increases in cALR@0.9. The corresponding changes in weighted F1 were small and representation dependent. The most reproducible consequence of MGAP was therefore anatomical: it altered the spatial distribution of class-related attribution more consistently than it altered diagnostic performance.

The segmentation comparison further clarifies the mechanism. The dual-bridge GAP model achieved slightly higher Dice than the MGAP counterpart but much lower ALR. Accurate prediction of pulmonary anatomy was therefore not equivalent to pulmonary concentration of the classifier's attribution. Auxiliary segmentation can establish that a representation contains sufficient information to reconstruct the lungs; it does not by itself establish that the final classifier uses that anatomy preferentially. Connecting the predicted soft mask directly to normalized aggregation makes this relationship operational rather than implicit.

The interpretation should nevertheless remain bounded. Because Grad-CAM for the proposed model is computed on the mask-influenced classifier representation, higher pulmonary containment is partly an expected consequence of the intervention being measured. The controlled GAP comparison reduces architectural confounding, but it does not convert Grad-CAM into a causal explanation. Future sensitivity analyses should therefore include matched pre-mask/pre-pooling attribution and perturbation-based faithfulness tests.

\subsection{External domain shift reveals multidimensional reliability}

RQ3 is addressed by the locked Shenzhen-to-Montgomery experiment. For DBCA-SegNet-MGAP, ROC-AUC decreased less than weighted F1, while ECE increased markedly. The source-domain probability scale and fixed threshold therefore transferred less successfully than ranking discrimination. This pattern is consistent with the broader observation that external radiologic performance often degrades under cohort shift \cite{yu2022external} and with the distinction between discrimination and calibration emphasized in neural-network calibration research \cite{guo2017calibration}.

Pulmonary attribution containment showed yet another transfer pattern. The proposed model retained ALR of approximately 0.65 and cALR@0.9 near 0.99 on Montgomery even though threshold-dependent performance and calibration worsened. These results reinforce that external robustness is not a single scalar property. A model may preserve ranking ability, lose calibration, and retain anatomically concentrated attribution to different degrees.

\subsection{Strong pulmonary containment is not sufficient for robust transfer}

The lightweight MGAP result is an important negative control. DBCA-Light + MGAP retained high Montgomery ALR and cALR@0.9 but its external weighted F1 fell sharply relative to DBCA-Light + GAP. Thus, stronger pulmonary containment alone did not guarantee that the model had learned disease features that generalized across cohorts.

This observation prevents overinterpretation of the anatomical metric. Whole-lung containment can reduce or expose extrapulmonary reliance, but the pulmonary field itself contains cohort-specific texture, acquisition effects, projection differences, and non-pathological variation. A classifier can therefore be ``inside the lungs'' spatially while still relying on features that do not transfer diagnostically. Pulmonary containment should be interpreted as one reliability dimension rather than a surrogate for robustness or clinical validity.

The efficiency results add a practical trade-off. The lightweight configurations reduced parameter count by approximately 86.9\%, but the full framework achieved a more favorable combination of external discrimination and pulmonary containment. Conversely, the larger architecture did not dominate every internal classification metric. No evaluated configuration simultaneously optimized diagnostic performance, calibration, anatomical containment, external transfer, and computational efficiency.

% ============================================================
\section{Conclusion and Future Work}
\label{sec:conclusion}
% ============================================================

This study evaluated CXR classification from a beyond-accuracy perspective by treating diagnostic performance, pulmonary attribution containment, calibration, and external robustness as related but distinct properties. Across three training seeds, DBCA-SegNet-MGAP achieved weighted F1 of $0.9615\pm0.0015$ and macro ROC-AUC of $0.9906\pm0.0007$ on the four-class COVID-19 task. More importantly, in the complete dual-BCAB representation, replacing GAP with MGAP increased ALR from $0.3878\pm0.0098$ to $0.7086\pm0.0104$ and cALR@0.9 from $0.5265\pm0.0101$ to $0.9905\pm0.0018$, while weighted F1 remained essentially unchanged. This separates accurate prediction of pulmonary anatomy from concentrating classifier attribution within that anatomy.

The locked Shenzhen-to-Montgomery experiment further showed that reliability properties do not transfer equally. On Montgomery, the proposed model retained ROC-AUC of $0.9080\pm0.0043$ and strong pulmonary attribution containment, whereas weighted F1 decreased to $0.7528\pm0.0080$ and ECE increased to $0.1683\pm0.0055$. The lightweight MGAP configuration provided a complementary counterexample: strong pulmonary containment could coexist with substantial external diagnostic degradation. Pulmonary containment is therefore not sufficient evidence of robust generalization.

Overall, diagnostic competence, pulmonary attribution containment, segmentation quality, calibration, and external transfer should be assessed separately. ALR and cALR remain containment metrics: they do not establish lesion localization, causal reasoning, or clinical correctness. The proposed framework should therefore be viewed as a controlled reliability-oriented research framework rather than evidence of clinical deployment readiness.

Future work should prioritize broader multi-center and prospective external evaluation, patient-level partitioning where identifiers are available, expert lesion annotations, and perturbation-based faithfulness tests that distinguish whole-lung containment from pathology-specific evidence. Independent calibration cohorts and clinically motivated operating-point selection are also needed. Finally, model compression, lower-cost cross-backbone interaction, and reliability-aware knowledge distillation should be investigated to reduce computational demand while preserving diagnostic performance, lung-mask quality, calibration, and pulmonary attribution behavior.

% ============================================================
\section*{Acknowledgement}
% ============================================================
During the preparation of this work, the authors used generative AI and AI-assisted technologies to improve grammar, readability, and formatting. After using these tools, the authors carefully reviewed and edited the content and take full responsibility for the contents of the published article.

% ============================================================
\section*{Data Availability}
% ============================================================
The COVID-19 Radiography Database, Shenzhen Tuberculosis Dataset, and Montgomery County Chest X-ray Dataset used in this study are publicly available from their respective cited repositories and original publications.

% ============================================================
\section*{Code Availability}
% ============================================================
All source code, training scripts, and evaluation notebooks are publicly available at \url{https://github.com/Abdullah-229/BeyondAccuracy}.

% ============================================================
\section*{Declaration of Competing Interest}
% ============================================================
The authors declare that there are no known conflicting interests or personal relationships which might have seemed to affect their research presented in this paper.

\bibliographystyle{elsarticle-num}
\bibliography{Main}

\end{document}